# Extended Active Learning Method

Ali Akbar Kiaei[1]      Saeed Bagheri Shouraki[2]      Seyed Hossein Khasteh[1]      Mahmoud Khademi[3]      Alireza Ghatreh Samani[2]

[1] Artificial Creatures Lab, Sharif University of Technology, Tehran, Iran
[2] Department of Electrical Engineering, Sharif University of Technology, Tehran, Iran
[3] Digital Signal Processing Lab, Sharif University of Technology, Tehran, Iran

kiaei@ce.sharif.edu      bagheri-s@sharif.edu      H_khasteh@ce.sharif.edu      Khademi@ce.sharif.edu      alirezasamany@ee.sharif.edu

*Abstract:* Active Learning Method (ALM) is a soft computing method which is used for modeling and control, based on fuzzy logic. Although ALM has shown that it acts well in dynamic environments, its operators cannot support it very well in complex situations due to losing data. Thus ALM can find better membership functions if more appropriate operators be chosen for it. This paper substituted two new operators instead of ALM original ones; which consequently renewed finding membership functions in a way superior to conventional ALM. This new method is called Extended Active Learning Method (EALM).

## 1 Introduction

Active Learning Method [1] is a powerful recursive fuzzy modeling without computational complexity. The main idea behind ALM is to break M.I.S.O. system into some S.I.S.O. subsystems and aggregate the behavior of subsystems to obtain the final output ([2], [3], [4]). This idea resembles the brain activity which stores the behavior of data instead of the exact values of them. Each S.I.S.O. subsystem is expressed as a data plane (called IDS plane) resulted from the projection of the gathered data on each input-output plane ([1], [5], [6]). Two types of information can be extracted from an IDS plane. First, the behavior of output respect to each input variable that is illustrated by a curve called narrow path. Second, the level of confidence for each input variable which is proportional to the reciprocal of variance of data around the narrow path.

Narrow paths are estimated by applying Ink Drop Spread (IDS) on data points and Center of Gravity (COG) on data planes ([12], [13], [14]). IDS and COG are two main operators of ALM ([1], [17]). But these operators have some problems. Actually when ALM breaks M.I.S.O system into different S.I.S.O systems, it loses some information. This is because ALM projects data into some X-Y planes ([1], [17]). And projecting an N dimensional space of inputs-outputs data, into (N-1) independent 2-d space can lead to losing data. This paper introduces/exploits two operators based on mathematical morphology, instead of original ALM operators. By this manner, ALM can cope with these problems.

As we will discuss in section 2, the operators of ALM are restated and it has been shown that one of its operators cannot support ALM very well. Two morphological algorithms are discussed in section 3 and it's proven that they satisfy Demorgans' law. Proposed fuzzy modeling method, EALM, with new operators is represented in section 4. Section 5 prepared various results and comparisons. And finally, Conclusion is declared in section 6.

## 2 Original operators of ALM

Flowchart of ALM is shown in Fig.1. The two operators, IDS and COG, are used to diffuse and fuse data in the space respectively ([16], [17]). IDS spreads the information in a plane and COG extracts the behavior of diffused data. These operators act as a fuzzy curve fitting method ([1], [5], [6]). They search for continuous possible paths by interpolating data points on data planes, in a fuzzy way.

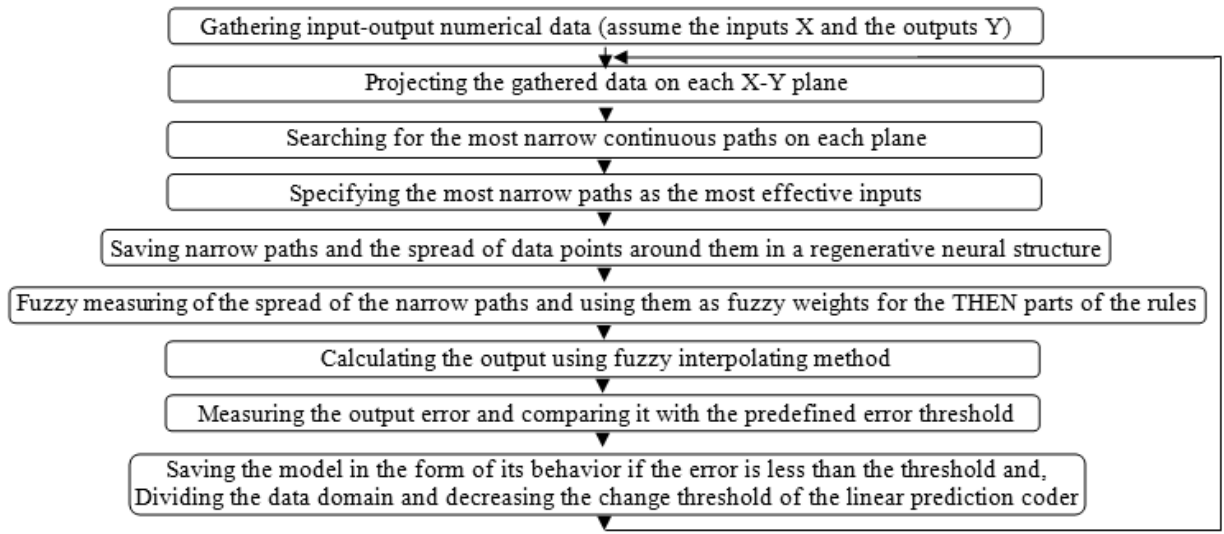

-Figure.1: proposed algorithm for Active Learning Method [1]

In the following, both operators, IDS and COG, are described. Then one example is shown in which they are used as ALM operators. Drawbacks of ALM regarding to their use are shown in the last part of this section.

### 2.1 IDS

IDS considers each data point on a data plane as a light source which has a cone shape illumination pattern [1]. This concept is illustrated in Fig.2, where this paper has used pyramids instead of cones. The projection of the process on the plane is called Ink Drop Spread. Pyramids can be considered as two dimensional (2-d) fuzzy membership functions which are centered on data points and can demonstrate the belief on data and their neighbors.

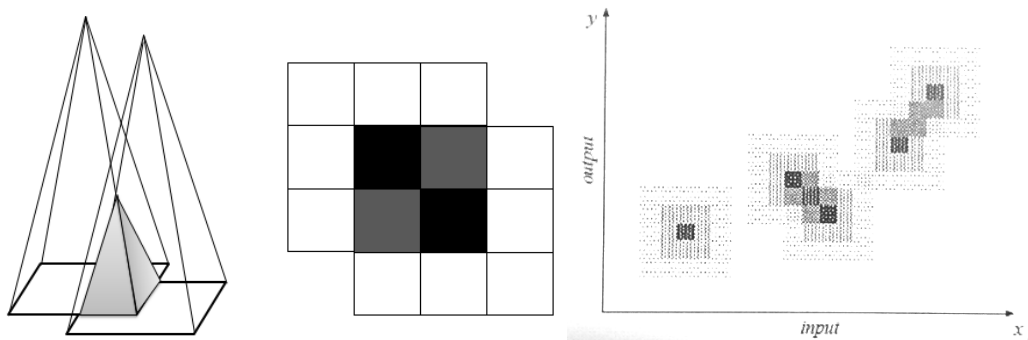

*-Figure 2: ink drop spread and fuzzy membership functions.*

The algorithm of IDS is applied on one data set by Murakami [12] as shown in Fig.3.

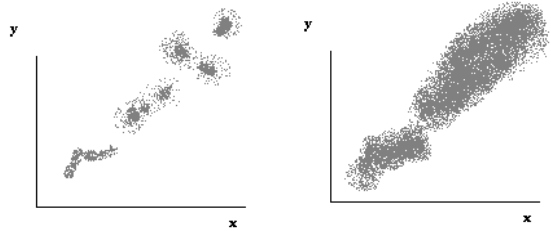

*-Figure 3: ink drop spread results for spread radius equal to 0.3 and 1 [12].*

## 2.2 Center of Gravity

The Center of Gravity method, tries to select a representative point in each column. This point can be obtained by calculating the weighted mean of all diffused data points in each column. Fig.4 shows the extracted path using COG on the plane shown in Fig.3

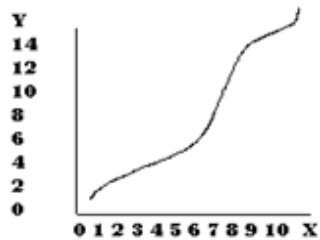

*-Figure 4: Extracted narrow path by Center of Gravity*

In Fig.5 you can see another example of COG. Each column has different data points; the representative point in each one is colored black.

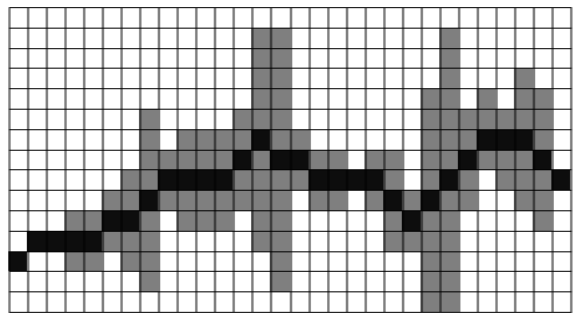

*-Figure 5: Center of Gravity in top view*

## 2.3 Modeling ALM by IDS and COG

In this part, the process of fuzzy modeling by ALM is shown, by giving an example. Assume that the input–output data have the relation given below:

$$x_1 = sin(t); \; x_2 = cos(t); \; y = sin(t); \; t \in [0, 10\pi]$$

The input output data is illustrated in Fig.6.

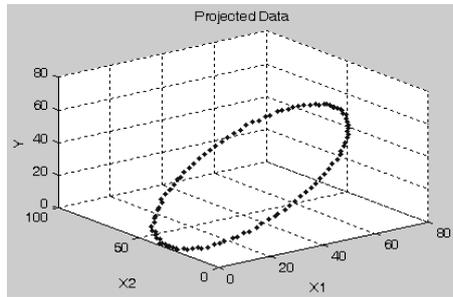

*-Figure 6: input output data in three-dimensional space*

At first, ALM projects this three dimensional space to two different two-dimensional input-output data sets, (x1, y), (x2, y), in which x1 and x2 are inputs and y is output. This is shown in Fig.7.

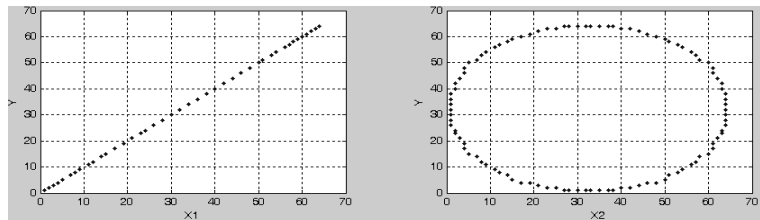

*-Figure7: projected data into different data planes*

In the next step, ALM applies IDS to each data plane, independently. This way, discrete points in the neighborhood of each other combine together as one continuous set. It is shown in Fig.8.

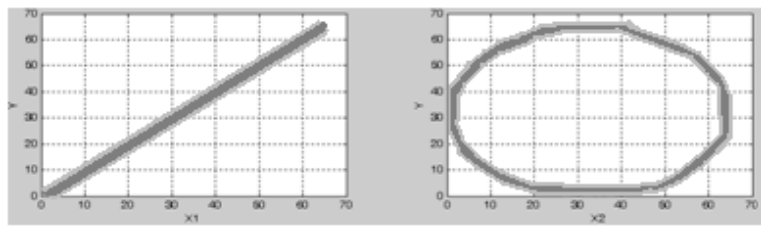

*-Figure 8: Applying IDS on each data plane*

Then ALM employs COG on each data plane. Actually, it selects one representative point in each column. These created narrow paths are representative data for each plane. Fig.9 shows these narrow paths by applying COG on Fig 8.

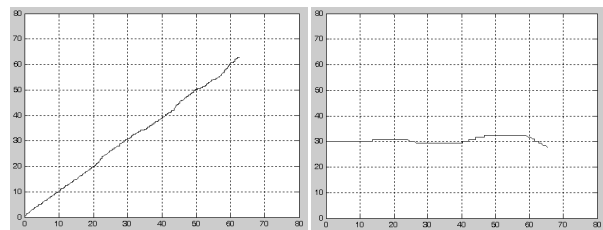

*-Figure 9: Applying COG after the IDS on each data plane*

Applying IDS and COG on the plane (x1, y) leads to good representative points for modeling, but applying the same on the plane (x2, y) destroys its structure. Originally the data on this plane had a circle shape structure, but the result has a line shape. In order to avoid such problems ALM defines "Truth" as a measure based on the accuracy of each representative point. It's formulated in this manner:

$$\text{Truth} = \exp[-\text{variance (data in that column)}]$$

So when the variance tends to zero, the Truth tends to one. In contrast, when the variance is very big, the Truth tends to zero. This variance is shown in Fig.10.

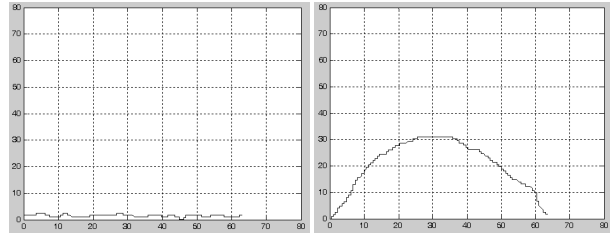

-Figure 10: the variance of data for each plane.

According to Fig.10, the accuracy of (x1, y) plane is good enough so that the first fuzzy rule can be made as given below:

> If $X_2$ is M  then  $Y = f_1(X_1)$
> $Truth = e^{-var(data)}$ ; $MF1 = M$

where "M" is the entire domain of x2, and MF1 is the membership function derived from M and Truth in each (x1, x2) [15]. The variance of data in the (x2, y)-plane is high. So the Truth value is low, and it cannot be considered a good fuzzy rule. In this stage ALM must find a point "t" in x1-axis to divide it into x1<t and x1>t, so that the data can be divided into upper and lower semi-circles. Assume that ALM divides (x1, y)- data plane into two areas, respectively named small and big. These areas are shown in Fig.11.

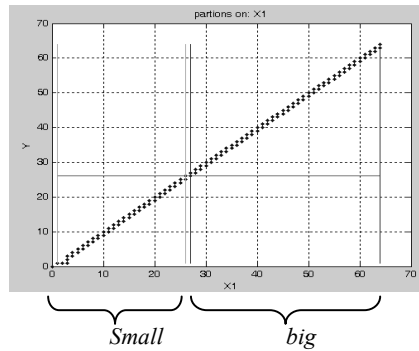

-Figure 11: two different areas for (x1, y)-plane

Consequently the (x2, y)-plane is divided into two data sets, based on data in small and big areas. These are upper and lower semi-circles. Now ALM can make two other fuzzy rules, because the accuracy of (x2, y)-plane is good enough so that these fuzzy rules can be made.

$$\text{If } X_1 \text{ is big} \quad \text{then } Y = f_1(X_2) \;; \qquad Truth = e^{-var(data)}$$
$$\text{If } X_1 \text{ is small} \quad \text{then } Y = f_3(X_2) \;; \qquad Truth = e^{-var(data)}$$

These rules are shown in Fig.12.

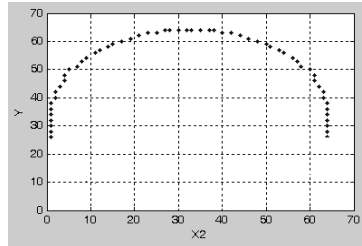

*If X1 is big then Y =*

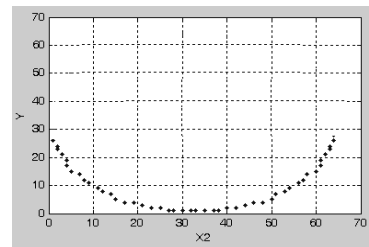

*If X1 is small then Y =*

*-Figure 12: divided (x2, y)-plane into two different planes*

### 2.4 The drawbacks

In this part three questions were asked. These questions can explain the disadvantages of ALM in complex situations.

First question: In the last example, when representatives were calculated, the structure of $(x_2, y)$- data plane was destroyed. Actually a circle-shape of structure is transformed to a line-shape structure of data. Does not this structure alteration defect the modeling? It has been shown that COG decreases the 2-dimensional space into a gathering of some 1-dimentional spaces side by side [1]. This decreasing can lead COG to lose some information. Actually COG tries to make a function from data in the plane while in general data may not present a function at all. So this operator cannot make a good representative for data necessarily. Therefore it should be replaced with a useful operator.

Second question: in the last example, since the variance of data on $(x_2, y)$-plane was high, the $(x_2, y)$-plane had to be divided into two areas. To reach this aim, ALM had to define two membership functions on $(x_1, y)$-plane, named small and big. How can ALM find these two membership functions? Although some ideas work fairly in special situations, ALM has not any accurate tool to achieve this aim in general. In the Extended Active Learning Method (EALM) the best membership functions can be made, so the number of fuzzy rules decreases and the accuracy of them increases.

Third question: ALM projects an N-dimensional space into different input-output planes, but it does not pay attention to any input-input plane. Are these planes completely useless? Do they convey no useful information? Actually ALM assumes that all inputs are independent of each other. But this makes ALM become susceptible to data rotation. For example, for one data ALM may make 16 fuzzy rules with an acceptable proficiency. 5 to 10 degree rotation of this data in the space may lead to making hundreds of fuzzy rules to reach that proficiency again.

These questions show that new operators should be replaced with ALM original operators and a new method should be introduced for exploiting them, so that the fuzzy modeling by ALM can be promoted.

In the next section, two new operators are introduced.

## 3 Two basic Morphological Algorithms

This section expresses two Algorithms that act like ALM operators, and it is shown that they are dual of each other.

## 3.1 Thinning

The Thinning [7] of one set, like A, by one structure element, like B, which is shown as "$A \otimes B$", is defined as below:

$$A \otimes B = A - (A \circledast B) = A \cap (A \circledast B)^c \quad 3.1.1$$

In which the Hit or Miss Transform (HOM) ([7], [8]) is defined:

$$A \circledast B = (A \ominus B1) \cap (A^c \ominus B2)$$

And Erosion [7] is defined:

$$A \ominus B = \{ c \in Z^2 | c + b \in A, \forall b \in B\}$$

Where B = (B1, B2). ([7], [9])

An applicable phrase for Thinning is based on chain of structuring elements [7]:

$$\{B\} = \{B^1, B^2, \ldots, B^n\} \quad 3.1.2$$

Where $B^i$ is a spun version of $B^{i-1}$. By this way, Thinning can be redefined in this manner:

$$A \otimes \{B\} = ((\ldots (A \otimes B^1) \otimes B^2) \ldots) \otimes B^n) \quad 3.1.3$$

First A is thinned by $B^1$, and then the result is thinned by $B^2$, and so forth. At last A is thinned by $B^n$. The whole action is repeated until no change happens. ([7], [10])

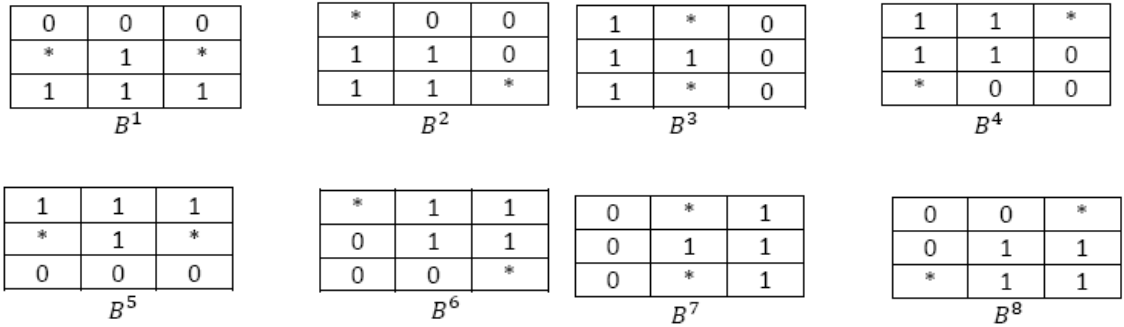

-Figure 13: Thinning structure elements [7]

## 3.2 Thickening

The Thickening ([7], [11]) of one set, like A, by one structure element, like B, which is shown as "$A \odot B$", is defined as below:

$$A \odot B = A \cup (A \circledast B) \quad 3.2.1$$

Thickening can be shown as sequential operations:

$$\{B\} = \{B^1, B^2, \ldots, B^n\} \quad 3.2.2$$

$$A \otimes \{B\} = ((\ldots (A \otimes B^1) \otimes B^2) \ldots) \otimes B^n) \quad 3.2.3$$

The structuring elements used for thickening have the same form as those shown in Fig.14 in connection with Thinning, but with all 1's and 0's interchanged[7].

| 1 | 1 | 1 |
|---|---|---|
| * | 0 | * |
| 0 | 0 | 0 |

$B^1$

| * | 1 | 1 |
|---|---|---|
| 0 | 0 | 1 |
| 0 | 0 | * |

$B^2$

| 0 | * | 1 |
|---|---|---|
| 0 | 0 | 1 |
| 0 | * | 1 |

$B^3$

| 0 | 0 | * |
|---|---|---|
| 0 | 0 | 1 |
| * | 1 | 1 |

$B^4$

| 0 | 0 | 0 |
|---|---|---|
| * | 0 | * |
| 1 | 1 | 1 |

$B^5$

| * | 0 | 0 |
|---|---|---|
| 1 | 0 | 0 |
| 1 | 1 | * |

$B^6$

| 1 | * | 0 |
|---|---|---|
| 1 | 0 | 0 |
| 1 | * | 0 |

$B^7$

| 1 | 1 | * |
|---|---|---|
| 1 | 0 | 0 |
| * | 0 | 0 |

$B^8$

-Figure 14: Thickening structure elements [7]

## 3.3 Thinning, Thickening and Demorgans law

Here, it is shown that Thickening is the dual of Thinning ([10], [11]).

The dual of Thinning is:

$$\left( \underbrace{A^c \otimes B^c}_{\text{①}} \right)^c = ? \qquad 3.3.1$$

To calculate this equation, compute ① regard to eq. 3.1.1. The ① is equal to:

$$① = A^c \otimes B^c = A^c \cap \left( \underbrace{A^c \circledast B^c}_{\text{②}} \right)^c \qquad 3.3.2$$

Based on the HOM definition [7] ② is rewritten as:

$$② = A^c \circledast B^c = (A^c \ominus B^c) \cap \left( A^{c^c} \ominus B^{c^c} \right) = (A^c \ominus B^c) \cap (A \ominus B) = (A \ominus B) \cap (A^c \ominus B^c) = A \circledast B$$

$$\Rightarrow A^c \circledast B^c = A \circledast B \qquad 3.3.3$$

So HOM of the elements are identical to their complement with the same action. Based on Thinning definition, it can be inferred that:

$$\Rightarrow ① = A^c \otimes B^c = A^c \cap (A^c \circledast B^c)^c \stackrel{3.3.3}{\iff} A^c \cap \underbrace{(A \circledast B)^c}_{\text{③}} \qquad 3.3.4$$

In the last equation, eq. 3.3.3 is used, to omit the extra relations. By expanding ③, inserting it into ①, and obtaining complement of ①, the dual of Thinning will be obtained.

$$③ = (A \circledast B)^c = [(A \ominus B) \cap (A^c \ominus B^c)]^c = \underbrace{(A \ominus B)^c}_{\text{④}} \cup (A^c \ominus B^c)^c \qquad 3.3.5$$

Thus, finding complement of HOM is simplified to the finding of complement of erosion, ④.

Moreover, the definition of dilation [7] is

$$A \oplus B = \{ x \mid (\hat{b})_x \cap A \neq \emptyset \}$$

Then the complement of erosion is rewritten respect to dilation:

$$④ = (A \ominus B)^c = \{ z \mid (B_z) \subseteq A \}^c = \{ z \mid B_z \cap A^c = \emptyset \}^c = \{ z \mid B_z \cap A^c \neq \emptyset \} = A^c \oplus \hat{B} \qquad 3.3.6$$

Inserting ④, ③, ② in their places, ① expanded as:

$$\Rightarrow \mathbb{1} = A^c \cap [(A^c \oplus \hat{B}) \cup (A \oplus \hat{B}^c)] = [A^c \cap (A^c \oplus \hat{B})] \cup [A^c \cap (A \oplus \hat{B}^c)]$$

And the dual of Thinning is

$$\Rightarrow (A^c \otimes B^c)^c = \mathbb{1}^c = \{[A^c \cap (A^c \oplus \hat{B})] \cup [A^c \cap (A \oplus \hat{B}^c)]\}^c$$
$$= [A^c \cap (A^c \oplus \hat{B})]^c \cap [A^c \cap (A \oplus \hat{B}^c)]^c = [A \cup (A^c \oplus \hat{B})^c] \cap [A \cup (A \oplus \hat{B}^c)^c]$$
$$= [A \cup (A \ominus B)] \cap [A \cup (A^c \ominus B^c)] = A \cup \{(A \ominus B) \cap (A^c \ominus B^c)\} = A \odot B$$

$$\Rightarrow (A^c \otimes B^c)^c = A \odot B \qquad 3.3.7$$

Thus, it is proven that Thickening is the dual of Thinning. By proving the contrary relation:

$(A^c \odot B^c)^c = (A \otimes B)$,

It is inferred that Thinning and Thickening are dual of each other.

By using the previous relations, it can be shown that:

$A^c \odot B^c = A^c \cup (A^c \circledast B^c) = A^c \cup (A \circledast B)$, so

$(A^c \odot B^c)^c = A \cap [(A \ominus B)^c \cup (A^c \ominus B^c)^c] = A - [(A \ominus B)^c \cup (A^c \ominus B^c)^c]^c = A - [(A \ominus B) \cap (A^c \ominus B^c)] = A - (A \circledast B) = A \otimes B \qquad 3.3.8$

Therefore, thinning and thickening are duals of each other ([7], [10], [11]).

## 4   New method of Fuzzy Modeling, EALM

This section explains the new method of fuzzy modeling, EALM (Extended Active Learning Method), by using Thinning and Thickening. This modeling method is detailed by an example.

Assume that input output data has a structure in three-dimensional space described below:

$$x_1 = sin(t); \; x_2 = cos(t); \; y = sin(t) + \cos(t); \; t \in [0, 10\pi]$$

The normalized and scaled data set is shown in Fig.15.

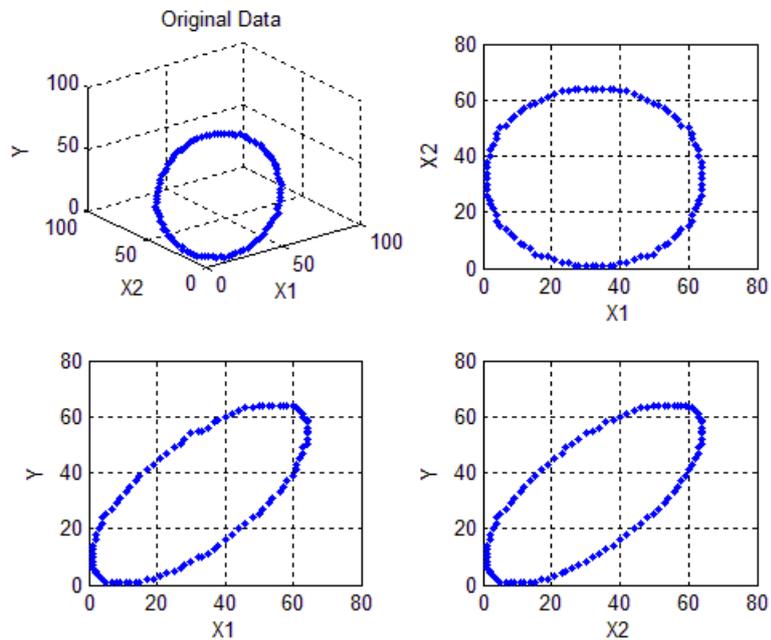

*-Figure 15: input-output data and its projection to the two dimensional planes*

At the first step, as you can see in Fig.15, EALM projects data into three different planes, $(x_1, y)$, $(x_2, y)$ and $(x_1, x_2)$. At the end part of this section you can see that how the last projection is useful.

At the second step, EALM uses the Thickening instead of IDS. You can see this in Fig.16.

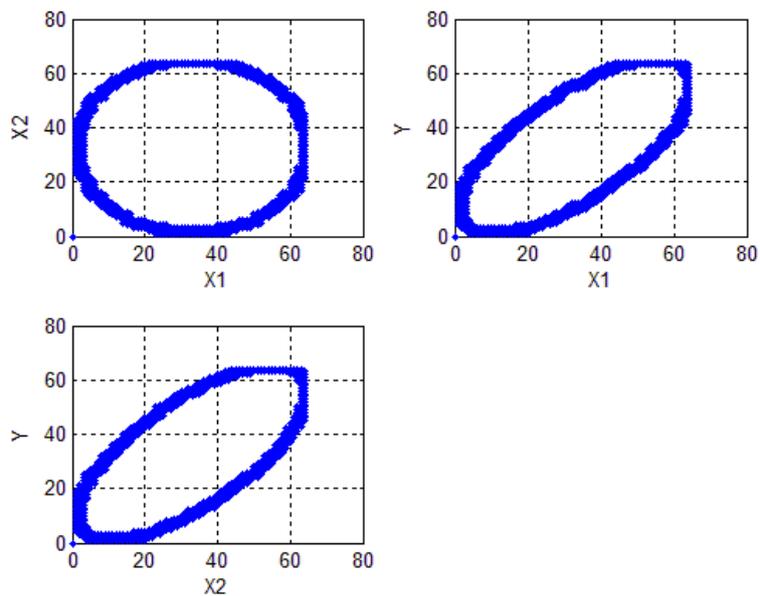

*-Figure 16: the effect of Thickening on each projected data*

The next step is finding appropriate delegates for the thickened data. EALM uses thinning algorithm instead of COG operator. This is shown in Fig.17. This process is done by EALM until no change occurs; therefore, the structure of data, which is called skeleton, can be preserved. After that noises are removed by some methods in morphology.

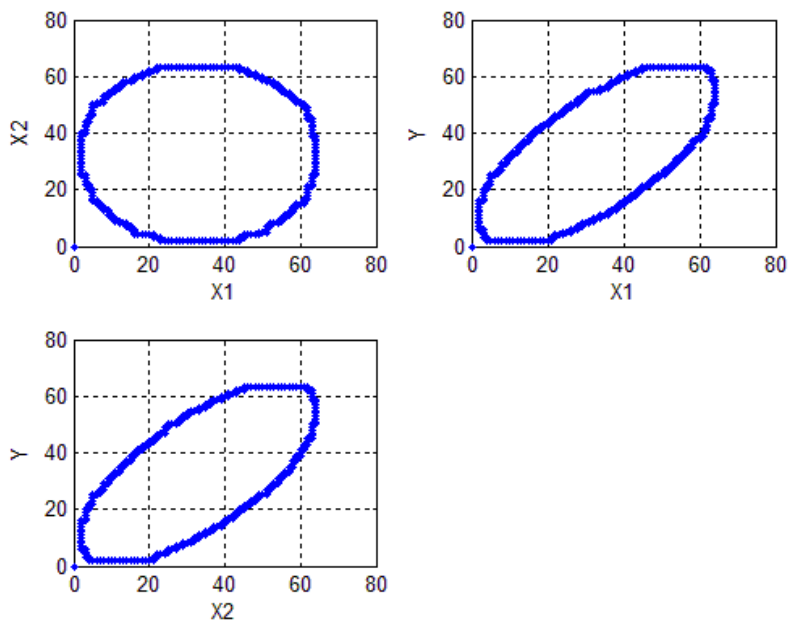

*-Figure 17: skeleton founded by thinning process again and again.*

As you can see in Fig.17, this way, unlike the COG, the structure of data did not change. Now EALM must find the best membership functions, but none of these delegates is a function. Thus it should divide these planes into semi-planes in order to declare powerful fuzzy rules for each produced part. To do so, EALM finds the best $y=y_0$ in input-output planes, such as planes $(x_1, y)$ and $(x_2, y)$, to make the maximum number of points like $(x, y_1)$ and $(x, y_2)$ which constraints $y_1<y_0, y_2>y_0$ hold. In this example, EALM found $y_0=30$, based on $(x_1, y)$-plane. Thus it separated other input-output planes into three areas. In first area, each point of data has a y that is less than or equal to $y_0$. In second area, each point has a y that is greater than $y_0$. Finally in third area, all other points in forms of $(x, y_1)$ and $(x, y_2)$ that satisfy the constraint $y_1<y_0, y_2>y_0$ exist. Actually the existence of last area implies that the inputs are dependent on each other. In the geometrical view, these areas can be independent of each other if the data rotates in a good manner. $y_0$ and these three areas are illustrated in Fig.18.

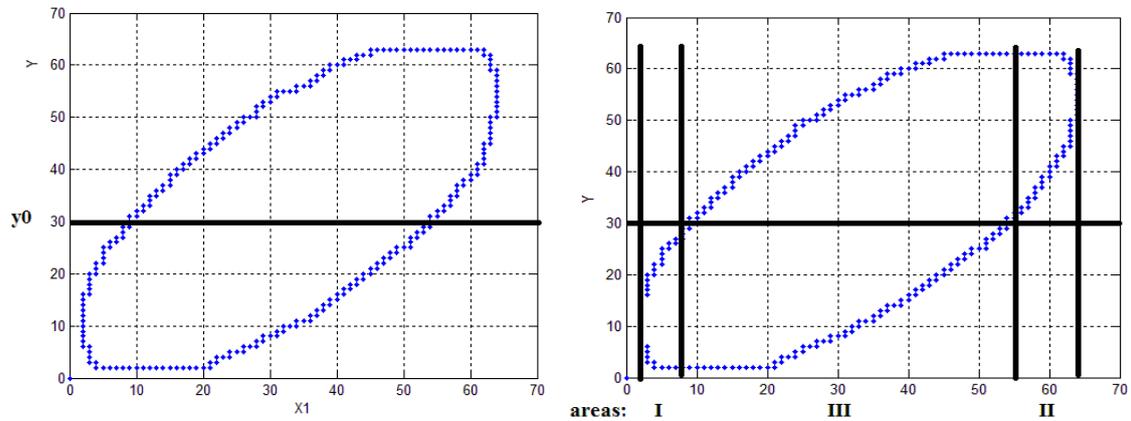

*-Figure 18: dividing data planes into semi-planes and separating them into areas*

As it can be seen in Fig.18, $(x_2, y)$-plane is separated into three areas. The first area, $(x_1, y)$-plane is limited to the lower part; and, the second area, $(x_1, y)$-plane is limited to the upper part. For the third area EALM classifies the data in this manner:

- Class 1: the data in the third area is in the form of $(x, y)$ which $y>y0$.
- Class 2: the data in the third area is in the form of $(x, y)$ which $y<y0$.

Then EALM projects these classes into related input-input plane, like $(x_1, x_2)$-plane in this example. A separate line can be calculated by one of pattern recognition methods. By this separate line, like $ax_1 + bx_2 + c = 0$ in this example, EALM separates these two classes and uses them as two membership functions. These classes are shown in Fig.19.

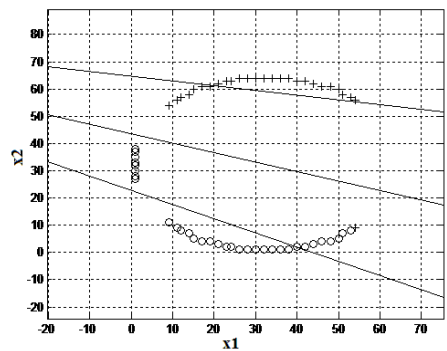

*-Figure 19: two classes of data in $(x_1, x_2)$-plane. "+" means class 1 and "o" means class 2.*

The separate line in Fig.19 can be shown as $ax_1 + bx_2 + c = 0$. So any point likes $(x_1, x_2)$ in the third area, which $ax_1 + bx_2 + c > 0$, is in the class 1. And any point likes $(x_1, x_2)$ in the third area, which $ax_1 + bx_2 + c < 0$, is in the class 2. Now if the amount of error is less than threshold, EALM introduces its membership functions like this:

- Membership Function **Small**: (X1 is area I) or $(ax1+ bx2+ c <0)$
- Membership Function **Big**: (X1 is area II) or $(ax1+ bx2+ c >0)$

Notice that EALM continues this dividing to make the amount of error less than a threshold. Because of this classification, EALM resists against the rotation and its membership functions are distributed properly in all data. The overview of EALM is shown in Fig.20. All steps in this figure are illustrated in the last example.

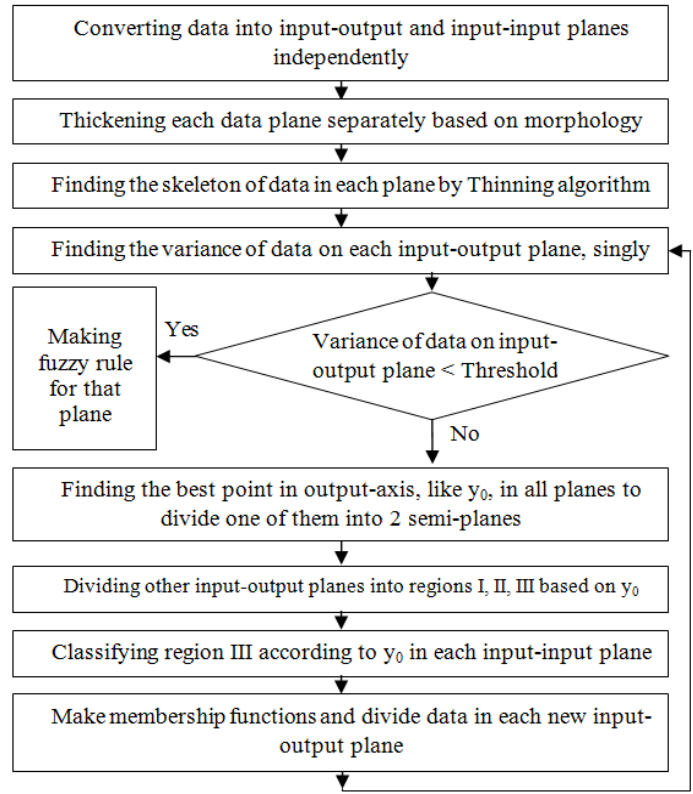

-Figure 20: overview of Extended Active Learning Method

# 5  Results

The first results are obtained on a sample data plane applying both of the original operators of ALM [1], IDS and COG which are executed in 5.1. Then the drawback of COG is shown from geometrical viewpoint. Thickening and Thinning are executed and it is shown that they do not have the disadvantage of COG. In addition, one more comparison between EALM and Murakami's results in ALM [1], is stated in 5.1. Two more comparisons between the results of Fuzzy modeling by EALM and ALM [12] are demonstrated in 5.2 as well.

## 5.1  Comparing the operators of ALM and EALM

Assume that the data points in the plane X-Y have a circular structure (or any structure which does not satisfy the structure of functions, $(\exists x_0: f(x_0) = y_1, f(x_0) = y_2$ and $y_1 \neq y_2)$. A sample data is shown in Fig.21.

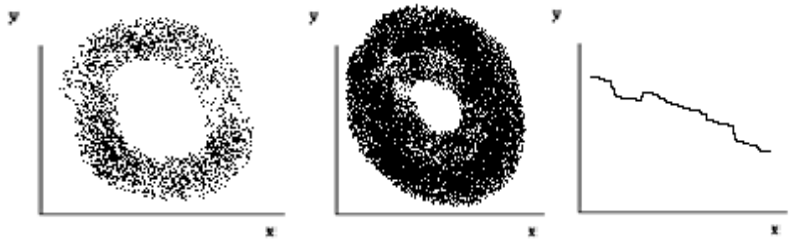

-Figure 21: sample data points with a structure like circle. Ink drop projection and extracted narrow path by Center of Gravity

By applying IDS and COG to this data plane, the strcture of data points are ruined. It is because in each column of data plane, center of gravity tries to select the average as a delegate of the column. It is shown in Fig.21. From the geometrical viewpoint, center of gravity reduces 2-d space to some 1-d spaces (called columns) which are located side by side [14]. It is because COG selects only one point, called delegate point [1] for each column in the data plane. Calculating delegates is independent of data in the neighbor columns. Since this space reduction causes destroying some information, the structure of original data points is not preserved necessarily.

Thinning keeps the skeleton ([7], [9]), thus the structure of data points is preserved. In the next step of ALM, data domain is divided to find narrow paths. Previously, ALM had to break narrow paths based on variance, because of lack of information. But now, EALM divides these narrow paths by the width of thickened data in any vertical lines. Thus EALM leads to an admissible answer faster. The fuzzy rules made by EALM are more accurate and numerically less than before. The results are shown in Fig.22.

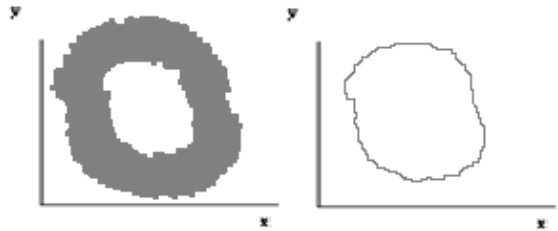

-Figure 22: Thickening projection and extracted narrow path by using Thinning.

As another example, assume that the projection of data points on plane X1-Y is shown in Fig 23.a. The result of performing IDS is shown in Fig.23.b and Fig.23.c. [1]

Fig.23.d shows the result of executing center of gravity. On the same data points Thickening is applied instead of IDS and Fig.24.a shows the result. Finally the narrow path, which is extracted by using Thinning, is illustrated in Fig.24.b. It has two new excess lines as compared to COG.

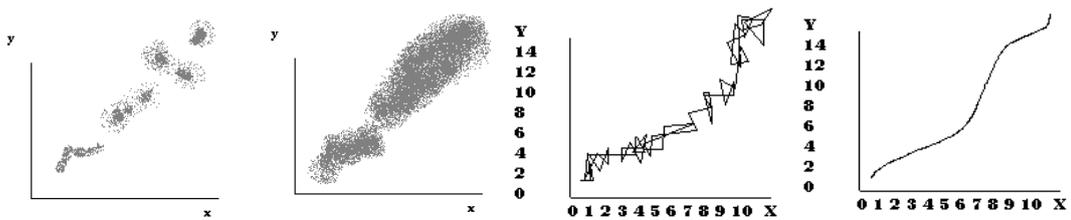

-Figure 23: main data points [1]; ink drop spread results (IDS) [1]; ink drop projection; extracted narrow path by Center of Gravity [1]

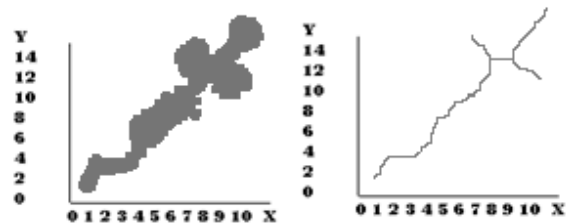

-Figure 24: Thickening projection and extracted narrow path by Thinning

The accuracy of narrow path and its details in comparison with IDS and COG is noticeable.

## 5.2 Comparing the fuzzy modeling by ALM and EALM

This part compares the result of modeling by ALM, based on Murakami [12], and the result of modeling by EALM.

The first data is followed by the function 5.2.1.

$$y = \sqrt{2\left(\frac{\sin x_1}{x_1}\right)^2 + 3\left(\frac{\sin x_2}{x_2}\right)^2}, \quad 1 \leq x_1, x_2 \leq 10 \qquad 5.2.1$$

This data is shown in Fig.25.a. Fig.25.b shows the result of modeling by ALM [12]. The result of modeling by EALM is shown in Fig.25.c. Like ALM, 450 data sets are used in modeling [12]. MSE and Correlation Coefficient are obtained from 1000 outputs of random inputs [12].

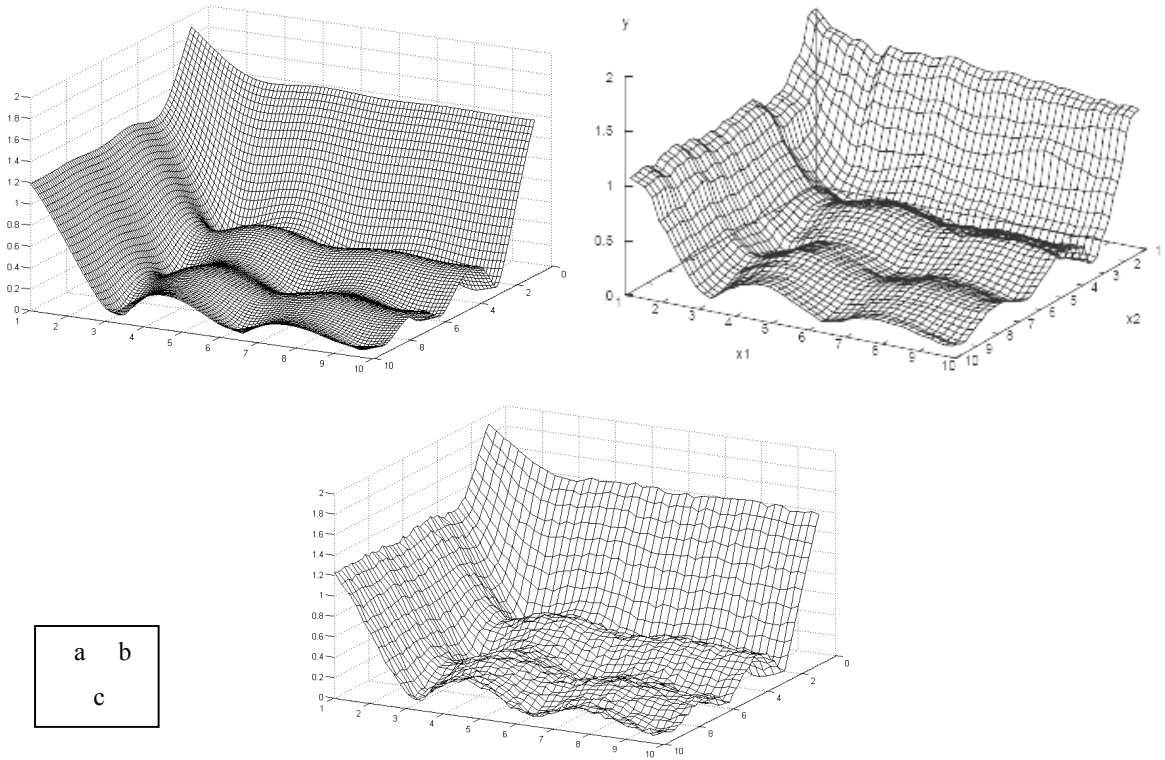

*Figure 25: the fuzzy modeling of data in Eq.5.2.1.   a) Main function; b) result of ALM [12]; c) result of EALM*

Mean square error and correlation coefficient for these two methods are mentioned in table 1.

| *Modeled Function 5.2.1* | *ALM* | *EALM* |
|---|---|---|
| *Mean Square Error* | 0.0020 | 0.0015 |

| **Correlation Coefficient** | 0.995 | 0.998 |
|---|---|---|

*Table 1: the comparison MSE and correlation coefficient between ALM [12] and EALM based on Eq.5.2.1*

The second data set is followed by the function 5.2.2.

$$y = (x_1 - 6 \sin x_2)^2, \ -10 \leq x_1 \leq 10, 0 \leq x_2 \leq 6 \qquad 5.2.2$$

This data is shown in Fig.26.a. Fig.26.b shows the result of modeling by ALM [12]. The result of modeling by EALM is shown in Fig.26.c. Like before, 450 data sets are used in modeling [12]. MSE and Correlation Coefficient are obtained from 1000 outputs of random inputs [12].

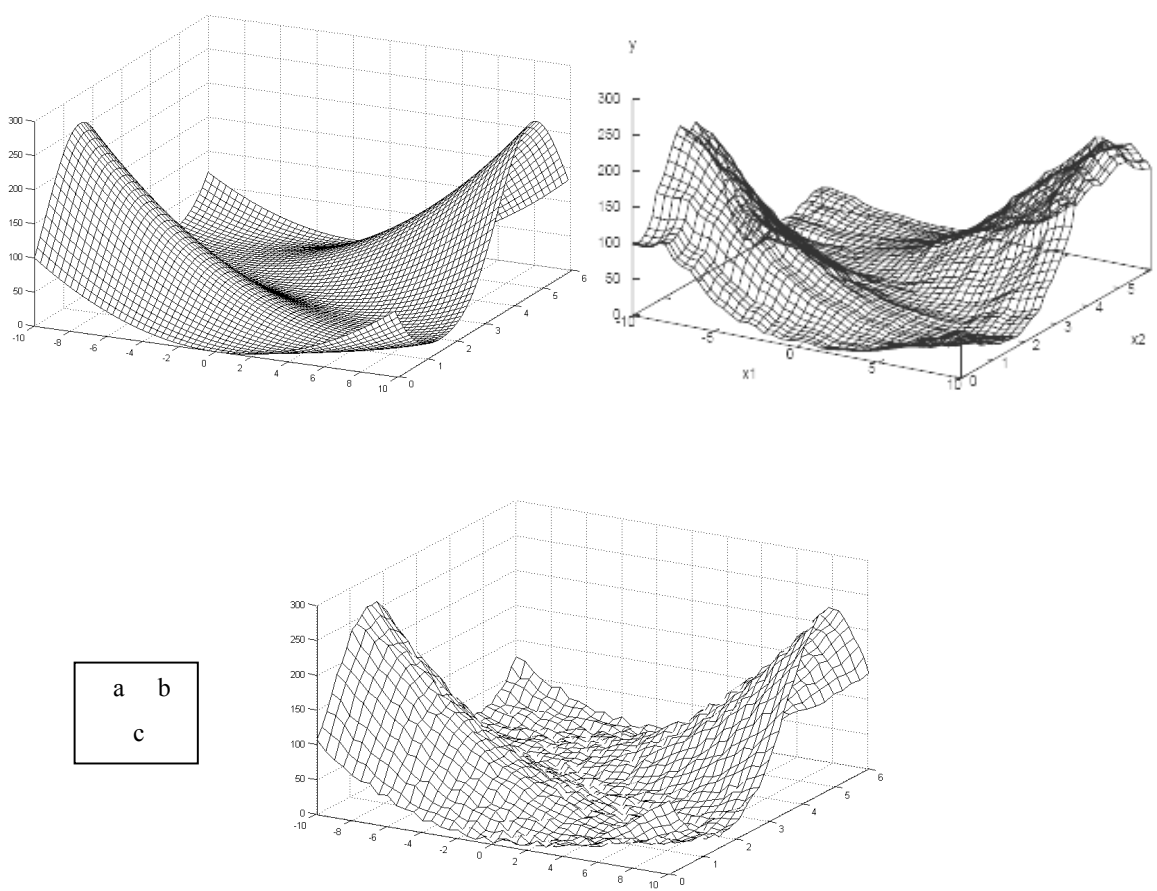

|     |     |
|-----|-----|
|  a  |  b  |
|     |  c  |

*-Figure 26: the fuzzy modeling of data in Eq.5.2.2.   a) Main function; b) results of ALM [12]; c) results of EALM*

Mean square error and correlation coefficient for these two methods are mentioned in the table 2.

| **Modeled Function 5.2.1** | *ALM* | *EALM* |
|---|---|---|
| *Mean Square Error* | *103* | *74* |
| *Correlation Coefficient* | *0.989* | *0.997* |

*Table 2: the comparison MSE and correlation coefficient between ALM [12] and EALM based on Eq.5.2.2*

# 6 Conclusion

In this paper we explained a fuzzy modeling, named as ALM. Then we showed that one of ALM operators loses some piece of information in data. So we introduced two morphological operators in order to be replaced by ALM original operators. Then the new method of finding membership functions was demonstrated. This new method of fuzzy modeling, called EALM, has shown some advantages over conventional ALM. Firstly, EALM does not lose data and consequently it finds the best membership functions. Secondly, it is not vulnerable to data rotation, so the minimum of fuzzy rules can be made. Finally, it can find few fuzzy rules with high accuracy due to its high ability of structure maintenance.

**Captions**

-Figure 1: proposed algorithm for Active Learning Method [1]
-Figure 2: ink drop spread and fuzzy membership functions.
-Figure 3: ink drop spread results for spread radius equal to 0.3 and 1 [12].
-Figure 4: Extracted narrow path by Center of Gravity
-Figure 5: Center of Gravity in top view
-Figure 6: input output data in three-dimensional space
-Figure 7: projected data into different data planes
-Figure 8: Applying IDS on each data plane
-Figure 9: Applying COG after the IDS on each data plane
-Figure 10: the variance of data for each plane.
-Figure 11: two different areas for (x1, y)-plane
-Figure 12: divided (x2, y)-plane into two different planes
-Figure 13: Thinning structure elements[7]
-Figure 14: Thickening structure elements[7]
-Figure 15: input-output data and its projection to the two dimensional planes
-Figure 16: the Thickening on each projected data
-Figure 17: skeleton founded by thinning process again and again.
-Figure 18: dividing data planes into semi-planes and separating them into areas
-Figure 19: two classes of data in (x1, x2)-plane. "+" means class 1 and "o" means class 2.
-Figure 20: overview of Extended Active Learning Method
-Figure 21: a sample data points with structure like circle. Ink drop projection and extracted narrow path by Center of Gravity
-Figure 22: Thickening projection and extracted narrow path by using Thinning.
-Figure 23: main data points [1]; ink drop spread results (IDS) [1]; ink drop projection; extracted narrow path by Center of Gravity [1]
-Figure 24: Thickening projection and extracted narrow path by Thinning
-Figure 25: the fuzzy modeling of data in Eq.5.2.1.   a) Main function; b) results of ALM [12]; c) results of EALM
-Figure 26: the fuzzy modeling of data in Eq.5.2.2.   a) Main function; b) results of ALM [12]; c) results of EALM